# Distributed Parallel Inference on Large Factor Graphs

**Joseph E. Gonzalez**
Carnegie Mellon University
jegonzal@cs.cmu.edu

**Yucheng Low**
Carnegie Mellon University
ylow@cs.cmu.edu

**Carlos Guestrin**
Carnegie Mellon University
guestrin@cs.cmu.edu

**David O'Hallaron**
Intel Research
david.ohallaron@intel.com

**Abstract**

As computer clusters become more common and the size of the problems encountered in the field of AI grows, there is an increasing demand for efficient parallel inference algorithms. We consider the problem of parallel inference on large factor graphs in the distributed memory setting of computer clusters. We develop a new efficient parallel inference algorithm, DBRSplash, which incorporates over-segmented graph partitioning, belief residual scheduling, and uniform work Splash operations. We empirically evaluate the DBRSplash algorithm on a 120 processor cluster and demonstrate linear to super-linear performance gains on large factor graph models.

## 1 INTRODUCTION

A computer cluster is a large collection of processors connected by a fast reliable communication network and configured to achieve a common task. Here we define a processor as a single processing element with a unique instruction counter[1]. Cluster computing confers both the obvious increase in computational throughput and memory capacity as well as the less obvious increase in memory bandwidth and cache capacity. With the availability of affordable commodity hardware and high performance networking, the AI community has increasing access to computer clusters. Unfortunately, many computationally intensive tasks in AI are not directly able to efficiently utilize cluster resources.

Work by Newman et al. [2007] and Asuncion et al. [2008] in parallel inference for latent topic models and by Paskin et al. [2005] and Funiak et al. [2006] in distributed inference for sensor networks adopt a message based asynchronous computation model to address the important task of distributed graphical model inference. However their approaches are specialized to particular models or settings. Alternatively, in Gonzalez et al. [2009] we explored the problem of general parallel inference in the multi-core shared memory setting. However, the shared memory model does not efficiently scale to large clusters, and the algorithm we proposed, ResidualSplash, makes scheduling assumptions that fail on large irregular models.

Here, we extend the ResidualSplash algorithm to large factor graphs in the distributed memory cluster setting and address several critical challenges to distributed parallel inference. We adopt the message passing computational model for cluster parallelism. In this model, the state of the algorithm is spread over $p$ processors which only exchange information by passing "messages." This differs from the shared memory setting of Gonzalez et al. [2009], where every processor has direct access to all available memory.

While the message passing model requires efficient work partitioning, and distributed reasoning, it also introduces several key advantages over the multi-core shared memory setting. Because processors are no longer tightly coupled, it is easier to construct substantially larger clusters. Because each processor has its own memory and dedicated bus, clusters provide increased memory capacity, memory bandwidth, and cache efficiency, permitting *super-linear* performance gains. Increased access to memory and memory bandwidth is critical to the performance of AI algorithms which often operate on large data-sets and quickly saturate the system bus.

In this paper, we outline the key challenges to efficient large scale distributed inference and address these challenges through the DBRSplash algorithm. The key contributions of this paper are:

- A formalization of state partitioning as a weighted graph cut and an empirical analysis of an approximate cutting procedure which exploits over-partitioning to improve work-balance.
- A belief residual scheduling and a work balanced Splash for improved scheduling on irregular graphs.
- DBRSplash, a distributed inference algorithm which retains the ResidualSplash parallel optimality.
- An empirical evaluation of DBRSplash on a 120 node cluster demonstrating linear to *super-linear* performance scaling for large factor graphs.

---
[1] We treat each core on a multi-core computer as a separate processor.

## 2 BELIEF PROPAGATION

Many important probabilistic models may be represented by factorized distributions of the form:

$$\mathbf{P}(x_1, \ldots, x_n) \propto \prod_{\alpha \in \mathcal{C}} \psi_\alpha(\mathbf{x}_\alpha), \quad (2.1)$$

where the set of factors $\mathcal{F} = \{\psi_\alpha : \alpha \in \mathcal{C}\}$ correspond to un-normalized positive functions, $\psi_\alpha : \mathbf{x}_\alpha \to \mathbb{R}^+$ over subsets $\mathcal{C} = \{\alpha : \mathbf{X}_\alpha \subseteq \mathcal{X}\}$ of the random variables. Here, we focus on discrete random variables $X_i \in \{1, \ldots, A_i\}$ taking on some finite set of $A_i$ possible values.

Distributions of the form Eq. (2.1) are naturally represented as a Factor Graph $G = (\{\mathcal{X}, \mathcal{F}\}, E)$, where the vertices $V = \mathcal{X} \cup \mathcal{F}$ are the variables and factors, and the edges $E = \{\{\psi_j, X_i\} : X_i \in \mathbf{X_j}\}$ connect factors with the variables in their domain. Factor graphs provide a convenient representation of the dependencies between variables which we will later exploit to partition the distribution over processors. To simplify notation, we use $\psi_i, X_j \in V$ to refer to vertices when we wish to distinguish between factors and variables, and $i, j \in V$ otherwise. We define $\Gamma_i$ as the neighbors of $i$ in the factor graph.

Estimating marginal distributions is essential to learning and inference in factor graphs. While computing exact marginals is NP-hard in general, there are many popular approximate inference algorithms. Belief Propagation (BP), or the Sum-Product algorithm, is a commonly used approximate inference algorithm originally proposed by Pearl [1988]. In BP, "messages" (parameters), are iteratively computed along edges in the factor graph until convergence and then used to estimate marginals. The message sent from variable $X_i$ to factor $\psi_j$ along the edge $\{X_i, \psi_j\}$ is given in Eq. (2.2) and the message sent from factor $\psi_j$ to vertex $X_i$ along the edge $\{\psi_j, X_i\}$ is given in Eq. (2.3),

$$m_{X_i \to \psi_j}(x_i) \propto \prod_{k \in \Gamma_i \setminus j} m_{k \to i}(x_i) \quad (2.2)$$

$$m_{\psi_j \to X_i}(x_i) \propto \sum_{\mathbf{x_j} \setminus x_i} \psi_j(\mathbf{x_j}) \prod_{k \in \Gamma_j \setminus i} m_{k \to j}(x_k) \quad (2.3)$$

where $\sum_{\mathbf{x_j} \setminus x_i}$ is a sum over all assignments to $\mathbf{x_j}$ with $x_i$ restricted, and $\prod_{k \in \Gamma_j \setminus i}$ is a product over all neighbors of the vertex $\psi_j$ excluding variable $X_i$.

In **synchronous BP**, all vertices *simultaneously* compute their outbound messages at every iteration using the messages from the previous iteration. In **asynchronous BP**, messages are updated *sequentially* using the most recent messages. Typically, message are sent until the maximum change in messages is bounded by a small constant $\beta \geq 0$:

$$\max_{(i,j) \in E} \left\| m_{i \to j}^{(\text{new})} - m_{i \to j}^{(\text{old})} \right\|_1 \leq \beta. \quad (2.4)$$

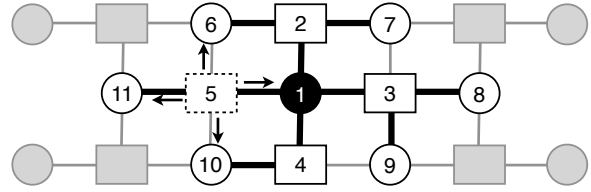

Figure 1: The Splash operation is being applied to the black vertex (1). The darker edges represent the BFS tree. The factors and variables are updated in the order $(11, 10, \ldots, 2, 1, 2, \ldots, 10, 11)$. The dotted factor ($5^{\text{th}}$ in the BFS) ordering is currently being updated resulting in new message represented by arrows.

The estimates of the marginal distributions are then,

$$\mathbf{P}(X_i = x_i) \approx b_{X_i}(x_i) \propto \prod_{j \in \Gamma_i} m_{j \to i}(x_i) \quad (2.5)$$

$$\mathbf{P}(\mathbf{X_i} = \mathbf{x_i}) \approx b_{\mathbf{X_i}}(\mathbf{x_i}) \propto \psi_i(\mathbf{x_i}) \prod_{j \in \Gamma_i} m_{j \to i}(x_j).$$

While BP is guaranteed to converge to the exact marginals in acyclic graphs, there are few guarantees for convergence or correctness in general graphs. Nonetheless, BP on cyclic graphs is used extensively with great success as an approximate inference algorithm [McEliece et al., 1998, Sun et al., 2003, Yedidia et al., 2003, Yanover and Weiss, 2002].

The "message" passing formulation and embarrassingly parallel synchronous update schedule suggests that BP is an embarrassingly parallel algorithm. However, in Gonzalez et al. [2009] we showed that *efficient* parallel inference in the shared memory setting is limited by the sequential dependencies among messages. Furthermore, the natural synchronous parallel scheduling can be asymptotically slower than the optimal parallel asynchronous scheduling. We provided a general asynchronous parallel algorithm, `ResidualSplash`, for the shared memory setting and demonstrated its optimality in the sequentially limiting case of chain graphical models.

### 2.1 THE RESIDUAL SPLASH ALGORITHM

Here, we briefly review the key points of the `ResidualSplash` algorithm which we extend in later sections. The Splash procedure, shown in Fig. 1, generalizes the optimal forward-backward sequential update ordering used in acyclic graphical models. When applied to a vertex $v$ in the factor graph, the Splash procedure first constructs a fixed volume breadth first search (BFS) ordering rooted at $v$. Then, starting at the leaves, vertices are sequentially updated until the root is reached and then the process is reversed. When a vertex is updated, all outbound messages from a vertex are recomputed using the current inbound messages. In an acyclic subgraph, the Splash procedure is equivalent to running forward-backward BP.

The `ResidualSplash` algorithm applies a variation of the residual scheduling heuristic proposed by Elidan et al. [2006] to determine the Splash ordering, prune the Splash BFS, and assess convergence. In particular the `ResidualSplash` algorithm assigns residual,

$$r_j = \max_{i \in \Gamma_j} \left\| m_{i \to j}^{\text{new}} - m_{i \to j}^{\text{old}} \right\|_1 \quad (2.6)$$

to each vertex. The value $r_j$ can be loosely interpreted as a measure of the value of updating vertex $j$. If $r_j = 0$ then updating a vertex will waste processor cycles as outgoing messages will not change. The `ResidualSplash` algorithm repeatedly runs the Splash procedure on the vertex with highest residual. When constructing the BFS, the Splash procedure stops searching along a branch if it reaches a vertex with residual less than the termination threshold $\beta$. Finally, `ResidualSplash` terminates when the highest residual is less than $\beta$.

## 3 DISTRIBUTING STATE

In this section, we address the challenges associated with distributing the state of the `ResidualSplash` algorithm over $p$ processors. In the shared memory setting, each processor has efficient direct access to all data structures and memory, permitting a single shared scheduling queue and message set. Consequently, the time and resources required to access the scheduling queue and update messages are symmetric for all processors. Conversely, in the distributed memory setting, access times are not symmetric. Instead, each processor can only directly access its local memory and must pass messages to communicate with other processors.

### 3.1 FACTOR GRAPH AND MESSAGES

We begin by partitioning the factor graph and messages. To maximize throughput and hide network latency, we must minimize communication and ensure that the data needed for message computations is locally available. We define a partitioning of the factor graph over $p$ processors as a set $\mathcal{B} = \{B_1, ..., B_p\}$ of disjoint sets of vertices $B_k \subseteq V$ such that $\cup_{k=1}^{p} B_k = V$. Given a partitioning $\mathcal{B}$ we assign all the factor data associated with $\psi_i \in B_k$ to the $k^{\text{th}}$ processor. Finally, we store each message on the processor containing the destination vertex.

Each vertex update is therefore a local procedure. For instance, if vertex $i$ is updated, the processor owning vertex $i$ can read factors and all incoming messages without communication. To maintain the locality invariant, after the new outgoing message is computed, it is transmitted to the processor owning the message.

By imposing the above locality constraints, we define the storage, computation, and communication responsibilities of each processor under a particular partitioning. Therefore, we can frame the minimum communication load balancing objective in terms of a graph partitioning, which is a classical problem in high performance computing. We formally define the graph partitioning problem as:

$$\min_{\mathcal{B}} \sum_{B \in \mathcal{B}} \sum_{(i \in B, j \notin B) \in E} w_{ij} \quad (3.1)$$

$$\text{subj. to:} \quad \forall B \in \mathcal{B} \quad \sum_{i \in B} w_i \leq \frac{\gamma}{p} \sum_{v \in v} w_v \quad (3.2)$$

where $w_{ij}$ is the communication cost of the edge between vertex $i$ and vertex $j$, $w_i$ is the total computation associated with the vertex $i$, and $\gamma \geq 1$ is the balance coefficient. The objective in Eq. (3.1) minimizes communication while for small $\gamma$, the constraint in Eq. (3.2) ensures work balance.

To define the communication and computation costs we introduce $U_i$ the total number of updates to vertex $i$. We define the communication cost as $w_{ij} = (U_i + U_j) \times (\min(|\mathcal{A}_i|, |\mathcal{A}_j|) + C_{\text{comm}})$ the total number of times a message is sent across the edge $(i, j)$ times the size of the message plus the fixed header cost $C_{\text{comm}}$. We define the work associated with each variable as $U_i \times |\Gamma_i| \times |\mathcal{A}_i|$, the number of updates, times the number of neighbors, times the size of that variable. Similarly, we define the work associated with each factor as $U_i \times |\Gamma_i| \times \prod_{j \in \Gamma_i} |\mathcal{A}_j|$, the number of updates. times the number of neighbors. times the size of the factor. While the exponential dependence on degree may suggest factors are more costly, their degree is usually small compared to variables.

Unfortunately, obtaining an optimal partitioning or near optimal partitioning is $NP$-Hard in general and the best approximation algorithms are generally slow. Fortunately, the there are several very fast heuristic approaches which typically produce reasonable partitions in time $O(|E|)$ linear in the number of edges. Here we use the collection of multi-level graph partitioning algorithms in the METIS [Karypis and Kumar, 1998] graph partitioning library. These algorithms, iteratively coarsen the underlying graph, apply high quality partitioning techniques to the small coarsened graph, and then iteratively refine the coarsened graph to obtain a high quality partitioning of the original graph. While there are no theoretical guarantees, these algorithms have been shown to perform well in practice and are commonly used in the parallel computing setting.

#### 3.1.1 Update Counts $U_i$

Unfortunately, due to dynamic scheduling, the update counts $U_i$ for each vertex depend on the evidence, graph structure, and progress towards convergence, and are not known before running `ResidualSplash`. In practice we find that the `ResidualSplash` algorithm updates vertices in a highly non-uniform manner; a key property of the dynamic scheduling, which enables more frequent updates of slower converging messages.

To illustrate the difficulty involved in estimating the update counts for each vertex, we introduce the synthetic denoising task. The input, shown in Fig. 2(a), is a grayscale image with independent Gaussian noise $N(0, \sigma^2)$ added to each pixel. The factor graph (Fig. 2(b)) corresponds to the pairwise grid Markov Random Field constructed by introducing a latent random variable for each pixel and connecting neighboring variables by factors that encode a similarity preference. The synthetic image was constructed to have a nonuniform update pattern (Fig. 2(d)) by making the top half more irregular than the bottom half. The distribution of vertex update frequencies (Fig. 2(f)) for the denoising task is nonuniform with a few vertices being updated orders of magnitude more frequently than the rest. The update patterns is temporally inconsistent frustrating attempts to estimate future update counts using past behavior. (Fig. 2(g)).

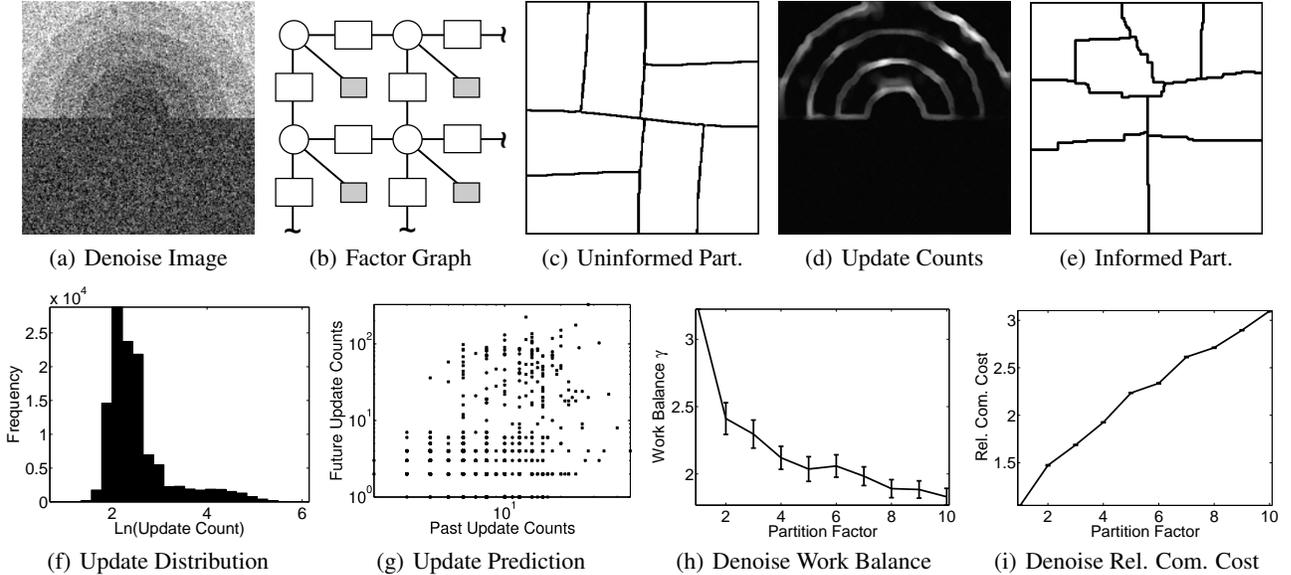

Figure 2: This figure illustrates the denoising problem, nonuniform update pattern of the `ResidualSplash` algorithm, the difference between uninformed and informed partitioning, and the impact of over-partitioning on uninformed partitions. **(a)** The synthetic noisy image. **(b)** Factor graph model for estimating for the denoising task. **(c)** Uniformed $U_i = 1$ balanced partitioning. **(d)** The update frequencies of each variable plotted in log intensity scale with brighter regions being more frequently updated. **(e)** The informed partitioning using the true update frequencies after running `ResidualSplash`. **(f)** The distribution of vertex update counts for an entire execution. **(g)** Update counts from first the half of execution plotted against update counts from the second half of the execution. **(h,i)** The effect of over-partitioning on the work balance and communication cost. For all points 30 trials with different random assignments are used and 95% confidence intervals are plotted. **(h)** The ratio of the size of the partition containing the most work, relative to the ideal size (smaller is better). **(i)** The communication cost relative to the informed partitioning.

### 3.1.2 Uninformed Partitioning

Surprisingly, in practice we find that an uninformed cut obtained by setting the number of updates to a constant (i.e., $U_i = 1$) achieves partitions with comparable communication cost and work balance as those obtained when using the true update counts. In Table 1 we construct uninformed $p = 120$ partitions $\hat{\mathcal{B}}$ with $\hat{U}_i = 1$ on several graphs and report the communication cost and balance

$$\text{Rel. Com. Cost} = \frac{\sum_{B \in \hat{\mathcal{B}}} \sum_{(u \in B, v \notin B) \in E} w_{uv}}{\sum_{B \in \mathcal{B}^*} \sum_{(u \in B, v \notin B) \in E} w_{uv}}$$

$$\text{Rel. Work Balance} = \frac{p}{\sum_{v \in V} w_v} \max_{B \in \hat{\mathcal{B}}} \sum_{v \in B} w_v$$

relative to the ideal cut $\mathcal{B}^*$ obtained using the true update counts $U_i$. We find that uninformed cuts have lower communication costs at the expense of increased imbalance. This discrepancy arises from the need to satisfy the balance requirement with respect to the true $U_i$ at the expense of a higher communication cost.

| Graph | Rel. Com. Cost | Rel. Work Balance |
|---|---|---|
| denoise | 0.980 | 3.44 |
| uw-systems | 0.877 | 1.837 |
| uw-languages | 1.114 | 2.213 |
| cora-1 | 1.039 | 1.801 |

Table 1: Comparison of communication cut cost and balance relative to the informed cut with known update frequencies and the uniformed cut with $U_i = 1$.

### 3.1.3 Over-Partitioning

Because uninformed partitions tend to have reduced communication cost and greater work imbalance relative to informed partitions, we propose over-partitioning to improve the overall work balance with a small increase in communication cost. When partitioning the graph with an uninformed cut a frequently updated subgraph may be placed within a single partition. To lessen the chance of such an event, we can over-partition the graph into $k \times p$ balanced partitions and then randomly redistribute the partitions to the original $p$ processors. By partitioning the graph more finely and randomly assigning regions to different processor, we more evenly distribute nonuniform update patterns improving the overall work balance. However, over-partitioning also increases the number of edges crossing the cut and therefore the communication cost. By over-partitioning in the denoise task we are able to improve the work balance (shown in Fig. 2(h)) at a small expense to the communication cost (shown in Fig. 2(i)).

Choosing the optimal over-partitioning factor $k$ is challenging and depends heavily on hardware, graph structure, and even factors. In situations where inference may be run repeatedly, standard search techniques may be used. We found that in practice when work balance is an issue small factors, e.g., $k = 5$ are typically sufficient. When using a recursive bisection partitioning algorithm where the true work split at each step is an unknown random variable, we can provide a theoretical bound on the ideal size of $k$. If at each split the work is divided into two parts of proportion $X$ and $1 - X$ where $\mathbf{E}[X] = \frac{1}{2}$ and $\mathbf{Var}[X] = \sigma^2$ ($\sigma \leq \frac{1}{2}$), Sanders [1994] shows that we can obtain work balance with high probability if we select $k$ at least $\Omega\left(p^{\left(\log\left(\frac{1}{\sigma+1/2}\right)\right)^{-1}}\right)$.

### 3.2 DISTRIBUTING THE PRIORITY QUEUE

The `ResidualSplash` algorithm relies on a shared global priority queue. However, in the cluster computing setting, a centralized ordering is inefficient. Instead, in our approach, each processor constructs a local priority queue and iteratively applies the Splash operation to the top element in its local queue. On each round, the globally highest residual vertex will be at the top of one of the local queues. Unfortunately, the remaining $p - 1$ highest vertices are not guaranteed to be at the top of the remaining queues and so we do not recover the original shared memory scheduling. However, any processor with vertices that have not yet converged, must eventually update those vertices and therefore can always make progress by updating the vertex at the top of its local queue. In Sec. 5.1 we show that the collection of local queues is sufficient to retain the original optimality properties of the `ResidualSplash` algorithm.

### 3.3 DISTRIBUTED TERMINATION

In the distributed setting where there is no synchronized common state, it is difficult to identify the globally largest element and stop the algorithm when it falls below the termination bound. This is the well studied distributed termination problem [Matocha and Camp, Mattern, 1987]. We implement a variation of of the algorithm described in Misra [1983] by defining a token ring over all the nodes, in which a marker is passed in one direction around the ring. The marker is advanced once the node owning the marker converges, halting execution. A node may resume execution if it receives a message that causes its maximum residual to exceed the termination threshold. Global termination is achieved when the token completes two cycles in which all nodes remain converged and the number of messages received equals the number of messages sent.

## 4 IMPROVED SCHEDULING

The fixed volume Splash operation and message based residual scheduling used in the `ResidualSplash` algorithm present several key challenges when scaling the algorithm to large factor graphs. In particular, both assume all vertices require the same amount of work to update. However, complex factors and variables that are involved in many factors often take much longer to update. Meanwhile, the message residual scheduling assumes that a significant change in one inbound message implies a significant change in the belief and outbound messages. Conversely, using message residuals as the convergence condition assumes that a small change in all inbound messages will induce only a small change in the belief and outbound messages. When the factor graph is large with high degree vertices this can result in an imbalanced convergence and an affinity for updating high degree vertices with little improvement in accuracy.

### 4.1 BALANCED SPLASH

When scheduling Splash operations the residual heuristic assigns a "value" to each vertex update which ignores the cost of computing the Splash. When the graph structure is regular, the size of each Splash and resulting costs are likely to be similar, enabling the residual scheduling to focus on minimizing the residual. However, when the cost of computing a Splash is vastly different, as is the case in large irregular graphs, the residual heuristic will fail to account for the cost. Consequently, the residual heuristic will skip relatively high residual vertices with low cost in favor of the highest residual vertex with much greater cost.

Furthermore, high degree, costly vertices are likely to be included in many BFS traversals and therefore updated disproportionately more often than other less heavily connected vertices. This problem is further frustrated in the cluster setting, by high degree vertices which are connected to vertices on many other processors, increasing network traffic substantially. We can resolve the imbalance in work by limiting the Splash size by the amount of work $W_{\max}$ (as defined in Sec. 3.1) rather than the number of vertices. Consequently, the scheduling heuristic can safely ignore the cost of each Splash.

### 4.2 NONUNIFORM CONVERGENCE

Using message residuals as the termination criterion leads to nonuniform convergence in beliefs. Small $\epsilon$ change to individual messages can combine at high degree vertices resulting in large changes in beliefs and asymmetric convergence. We demonstrate this behavior by considering a variable $X_i$ with $d = |\Gamma_i|$ incoming messages $\{m_1, \ldots, m_d\}$. Suppose all the incoming messages are changed to $\{m'_1, \ldots, m'_d\}$ such that the resulting residual is less than $\beta$ (i.e., $\forall k \, |m'_k - m_k|_1 \leq \beta$). Using the convergence criterion in Eq. (2.4) the messages have converged. However, the effective change in belief depends linearly on the degree, and therefore can be far from convergence.

Assume $\{m_1, \ldots, m_d\}$ are binary uniform messages. Then the belief at that variable is also uniform (i.e., $b_i = [\frac{1}{2}, \frac{1}{2}]$). If we then perturb the messages $m'_k(0) = \frac{1}{2} - \epsilon$ and $m'_k(1) = \frac{1}{2} + \epsilon$ by some small $\epsilon \leq \beta/2$ the new belief is:

$$b'_i(0) = \frac{\left(\frac{1}{2} - \epsilon\right)^d}{\left(\frac{1}{2} + \epsilon\right)^d + \left(\frac{1}{2} - \epsilon\right)^d}.$$

The $L_1$ belief residual due to the compounded $\epsilon$ change in each message is then:

$$|b'_i(0) - b_i(0)|_1 = \frac{1}{2} - \frac{\left(\frac{1}{2} - \epsilon\right)^d}{\left(\frac{1}{2} + \epsilon\right)^d + \left(\frac{1}{2} - \epsilon\right)^d}.$$

A $2^{\text{nd}}$ order Taylor expansion around $\epsilon = 0$ obtains:

$$|b'_i(0) - b_i(0)|_1 \approx d\epsilon + O(\epsilon^3).$$

Therefore, the change in belief varies linearly in the degree of the vertex enabling small $\epsilon$ message residuals to translate into large $d\epsilon$ belief residuals.

### 4.3 BELIEF RESIDUALS

The aim of BP is to estimate the marginal for each variable. However, `ResidualSplash` defines the scheduling and convergence using the change in messages rather than beliefs. In Sec. 4.2, we showed that small message changes do not imply small belief changes. Here we define a belief residual which addresses the problems associated with the message-centric approach.

A natural definition of the belief residuals analogous to the message residuals defined in Eq. (2.6) is

$$r_j = \left|\left|b_i^{\text{new}} - b_i^{\text{old}}\right|\right|_1 \qquad (4.1)$$

where $b_i^{\text{old}}$ is the belief at vertex $i$ the last time vertex $i$ was updated. Unfortunately, Eq. (4.1) has a surprising flaw that admits premature convergence on acyclic graphs with $\beta = 0$ under a specially constructed scheduling. We will demonstrate this failure scenario and present a natural solution which is also computationally desirable.

Without loss of generality we consider a chain MRF of 5 vertices with the binary factors $\psi_{X_i, X_{i+1}}(x_i, x_{i+1}) = I[x_i = x_{i+1}]$ and unary factors:

$$\psi_{X_1} = \left[\tfrac{1}{9}, 9\right] \quad \psi_{X_2} = \left[\tfrac{9}{10}, \tfrac{1}{10}\right] \quad \psi_{X_3} = \left[\tfrac{1}{2}, \tfrac{1}{2}\right]$$

$$\psi_{X_4} = \left[\tfrac{1}{10}, \tfrac{9}{10}\right] \quad \psi_{X_5} = \left[9, \tfrac{1}{9}\right]$$

We begin by initalizing all vertex residuals to infinity, and all messages to uniform distributions. Then we perform the following update sequence marked in black:

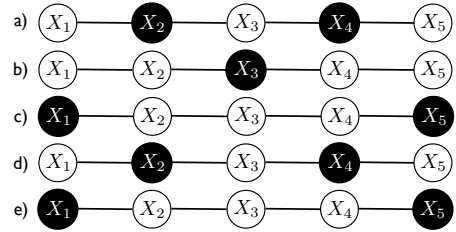

After stage (b), $X_3$ will have uniform belief and zero residual and $m_{2\to 3} = \psi_{X_2}$ and $m_{4\to 3} = \psi_{X_4}$. After stage (d), $m_{2\to 3}$ and $m_{4\to 3}$ will have swapped values. Therefore $X_3$ will continue to have uniform belief and zero residual. At this point $X_2$ and $X_4$ also have zero residual since they were just updated. Stage (e) clears the residuals on $X_1$ and $X_5$. The residuals on $X_2$ and $X_4$ remain zero since messages $m_{1\to 2}$ and $m_{5\to 4}$ haven't changed since state (c). By Eq. (4.1) with $\beta = 0$ we have converged prematurely since no sequence of messages connects $X_1$ and $X_5$. The use of the naive belief residual in Eq. (4.1) will therefore converge to an erroneous solution.

An alternative formulation of the belief residual which does not suffer from premature convergence is given by:

$$r_j^{(t)} \leftarrow r_j^{(t-1)} + \left|\left|b_i^{(t)} - b_i^{(t-1)}\right|\right|_1 \qquad (4.2)$$

$$b_i^{(t)}(x_i) \propto \frac{b_i^{(t-1)}(x_i) \, m_{i\to j}^{(t)}(x_i)}{m_{i\to j}^{(t-1)}(x_i)}. \qquad (4.3)$$

$b_i^{(t-1)}$ is the belief after incorporating the last message and $b_i^{(t)}$ is the belief after incorporating the new message. As each new message arrives, the belief can be efficiently recomputed using Eq. (4.3). Because Eq. (4.2) accumulates the change in belief with each new message, it will not lead to premature termination. Intuitively, it measures the cumulative effect of all message updates on the belief. Additionally, since Eq. (4.2) satisfies the triangle inequality, it is an upper bound on the total change in belief. This residual definition also has the advantage of not requiring previous versions of messages or beliefs to be stored.

## 5   THE DBRSPLASH ALGORITHM

We now present our Distributed Belief Residual Splash algorithm (`DBRSplash` shown in Alg. 1) which combines the ideas presented in earlier sections. The execution can be divided into two phases, setup and inference.

In the setup phase, in Line 1 we over-segment the input factor graph into $kp$ pieces using the METIS algorithms. Note that this could be accomplished in parallel using ParMETIS, however our implementation uses the sequential version for simplicity. Then in Line 2 we randomly assign $k$ pieces to each of the $p$ processors. In parallel each processor collects its factors and variables (Line 3). On

**Algorithm 1**: The DBRSplash Algorithm

1  $\mathcal{B}_{\text{temp}} \leftarrow$ OverSegment$(G, p, k)$;
2  $\mathcal{B} \leftarrow$ RandomAssign$(\mathcal{B}_{temp}, p)$;
   **forall** *Processors* $b \in \mathcal{B}$ **do in parallel**
3  $\quad$ Collect$(\mathcal{F}_b, \mathcal{X}_b)$;
4  $\quad$ Initialize$(Q)$;
5  $\quad$ **while** TokenRing$(Q, \beta)$ **do**
   $\qquad v \leftarrow$ Pop$(Q)$;
6  $\qquad$ FixedWorkSplash$(v, W_{\max}, \beta)$;
7  $\qquad$ RecvExternalMsgs$()$;
8  $\qquad$ **foreach** $u \in$ Local changed vertices **do**
9  $\qquad\quad$ Promote$(Q, ||\Delta b_v||_1)$;
10 $\qquad$ SendExternalMsgs$()$;
   $\qquad$ Push$(Q, v, 0)$;

Line 4 the priorities of each variable and factor are set to infinity to ensure that every vertex is updated at least once.

On Line 5 we evaluate the top residual with respect to the $\beta$ convergence criterion and check for termination in the token ring. On Line 6, a splash of total work $W_{\max}$ is applied to $v$. The fixed work Splash uses $\beta$ to prune subtrees that have sufficiently low belief residual. After completing the Splash all external messages from other processors are incorporated (Line 7). Any beliefs that changed during the Splash or after receiving external messages are promoted in the priority queue on Line 9. On Line 10, the external messages are transmitted across the network. Empirically, we find that accumulating external messages and transmitting only once every 10 loops reduces network overhead substantially and does not adversely affect convergence. The process repeats until termination at which point all beliefs are sent to the originating processor.

### 5.1 PRESERVING SPLASH CHAIN OPTIMALITY

In Gonzalez et al. [2009] we introduced the $\tau_\epsilon$ notation as a theoretical measure for the effective distance $\tau_\epsilon$ at which vertices are assumed to be almost independent. More formally, for all vertices $\tau_\epsilon$ is the minimum radius for which running belief propagation on the subgraph centered at that vertex yields beliefs at most $\epsilon$ away from beliefs obtained using the entire graph. By increasing the value of $\epsilon$, we decrease the sequential dependency structure, and increase the opportunity for parallelism. We showed that the ResidualSplash algorithm, when applied to chain graphs under the $\tau_\epsilon$ approximate inference setting, achieves the $\Omega(|V|/p + \tau_\epsilon)$ optimal lower bound. We now show that DBRSplash retains the optimality in the distributed setting.

**Theorem 5.1** (Splash Chain Optimality). *Given a chain graph with $n = |V|$ vertices and $p \leq n$ processes, the distributed DBRSplash algorithm with no over-segmentation, using a graph partitioning algorithm which returns connected partitions, and with work Splash size at least $2\sum_{v \in V} w_v/p$ will obtain a $\tau_\epsilon$-approximation in expected running time $O\left(\frac{|V|}{p} + \tau_\epsilon\right)$.*

*Proof of Theorem 5.1.* The proof is essentially identical to the method used in Gonzalez et al. [2009]. We assume that the chain graph is optimally sliced into $p$ connected pieces of $|V|/p$ vertices each. Since every vertex has at most 2 neighbors, the partition has at most $2|V|/p$ work. A Splash anywhere within each partition will therefore cover the entire partition, performing the complete "forward-backward" scheduling.

Because we send and receive all external messages after every splash, after $\lceil \frac{\tau_\epsilon}{|V|/p} \rceil$ iterations, every vertex will have received messages from vertices a distance of at least $\tau_\epsilon$ away. The runtime will therefore be:

$$\frac{2|V|}{p} \times \left\lceil \frac{\tau_\epsilon}{|V|/p} \right\rceil \leq \frac{2|V|}{p} + 2\tau_\epsilon$$

Since each processor only send 2 external messages per iteration (one from each end of the partition), communication therefore only adds a constant to the total runtime. □

## 6 EXPERIMENTS

We implemented an MPI based version of DBRSplash in C++ using MPICH2. The splash size, over-partitioning factor, and scheduling method (i.e., message based and belief based) were parametrized for comparison. We invoked the weighted *kmetis* partitioning routine from the METIS software library for graph partitioning. All partitions were computed in under 10 seconds. To ensure numerical stability and convergence, log-space message calculations and 0.6 damping were used. The convergence bound was set to $\beta = 10^{-5}$. Cluster experiments were compiled using GCC 4.2.4 and tested on a cluster of 15 64Bit Linux Blades with dual Quad-Core Intel Xeon 2.33GHZ (E5345) processors connected with Gigabit Ethernet.

We assessed the performance of DBRSplash on Markov Logic Networks (MLNs) [Domingos et al., 2008], a probabilistic extension to first-order logic obtained by attaching weights to logical clauses. We used *Alchemy* to compile several MLNs into factor graphs. We constructed MLNs from the UW-CSE relational data-set [Domingos, 2009] and present results for the smallest *uw-languages* MLN with 1078 variables and 26598 factors and the largest *uw-systems* MLN with 7951 variables and 406389 factors. These MLNs have varied degree distributions as seen in Fig. 3. The large *uw-systems* model illustrates the scaling potential while *uw-languages* illustrates the limitations of our algorithm on small models.

### 6.1 PARALLEL PERFORMANCE

The running time and speedup of DBRSplash were assessed on the *uw-systems* and *uw-languages* MLNs using various over-partitioning factors. In Fig. 4(a) and Fig. 4(b), `DBRSplash` achieves linear to super-linear running times and speedups up to 120 processors on the larger *uw-systems* MLN. The super-linear speedup may be attributed to increasing cache efficiency and memory bandwidth. Increasing the over-partitioning factor initially improves performance but performance gains are gradually attenuated by increased communication costs as more processors are used. Meanwhile, the much smaller *uw-languages* MLN only demonstrates linear to super-linear performance gains up to 20 processors (Fig. 5(a) and Fig. 5(b)). With a total running time under 10 seconds, there is insufficient work to efficiently use more than 20 processors.

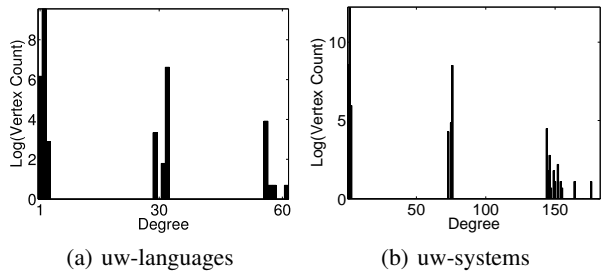

(a) uw-languages (b) uw-systems

Figure 3: The irregular degree distributions for the (a) *uw-languages* and (b) *uw-systems* MLNs.

### 6.2 OVER-PARTITIONING

To directly assess the impact of over-partitioning on work balance and network traffic we used the denoising task (introduced in Sec. 3.1.1) on a $500 \times 500$ image with 5 colors. Using 60 processors we tested several over-partition factors and plotted both instantaneous CPU usage (Fig. 6(a)) and the cumulative network traffic (Fig. 6(b)). Without over-partitioning, the computation is unbalanced resulting in a gradual decrease in the number of active processors. Increasing the over-partitioning factor decreases the running time and ensures that all processor remain active up to convergence. However, as suggested, over-partitioning increases network activity.

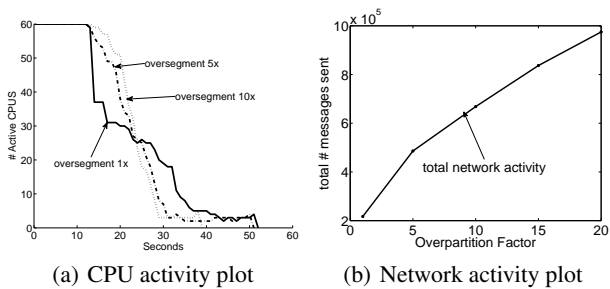

(a) CPU activity plot (b) Network activity plot

Figure 6: (a) Number of processors active for a 500x500 denoising image with 5 different colors, using a maximum of 60 processors. (b) Total number of network messages sent. Over-partitioning reduces running time, improves processor utilization, but increases network usage.

We conduct a similar analysis on both MLNs. The cpu usage Fig. 4(d) for the *uw-systems* MLN is consistent with results from the denoising task. Surprisingly, the network activity Fig. 4(e) for the *uw-systems* MLN after an initial increase shows a minor decrease with increased over-partitioning which we attribute to variability in partitioning and reduced running time. The cpu usage (Fig. 5(d)) for the smaller *uw-languages* MLN shows an increase in balance with increasing partitioning factor, but performance decreases going from over-partitioning factor of 5 to 10. This is due to the increased network activity (Fig. 5(e)) dominating the already short computation time.

### 6.3 ACCURACY ASSESSMENT

To assess the accuracy of `DBRSplash` belief estimates we compare with belief estimates obtained through Gibbs sampling. We generated chains of 125 thousand samples, dropped the first 25 thousand samples (burn-in), and then used remaining samples to estimate the true beliefs. We compared belief estimates by computing the $L_1$ difference averaged over all variables in the model. We found that repeated chains starting at different random states converged to beliefs that differed by less than 0.05 in average $L_1$ per variable.

In Fig. 4(c), we plot the accuracy of `DBRSplash` with 60 processors on *uw-systems* as function of the vertex updates. In Fig. 5(c), we do the same for *uw-languages* but using a single processor since the running time is too short. In both cases `DBRSplash` quickly achieves high accuracy. In Fig. 4(f), we plot the accuracy of `DBRSplash` on *uw-systems* as a function of the number of processors for a fixed running time of one minute. We see that using 20 processors substantially improves the running time but the return quickly diminishes. In Fig. 5(f) we do the same for *uw-languages*, but for a running time of one second since it converges rapidly. In this case, going beyond 20 processors decreases the accuracy, due to increased running time.

### 6.4 IMPROVED SCHEDULING

It is difficult to directly compare the convergence time of belief based scheduling and message based scheduling because they do not share the same convergence criterion. To provide a common basis for comparison, we assess the accuracy of the beliefs as discussed in Sec. 6.3 (Fig. 4(c), Fig. 5(c), Fig. 4(f) and Fig. 5(f)). These plots compare the accuracy of `DBRSplash` using Belief Residuals and Message Residuals. For *uw-systems*, the belief residuals achieve more rapid convergence to an accurate solution. The *uw-languages* MLN presents nearly identical accuracy convergence using both scheduling methods.

Additionally, we have found that for several MLNs, message based scheduling failed to converge while belief based scheduling converges consistently. One such MLN, *cora-1*, is characterized by extremely high degree variables (e.g., 59 variables with degree greater than 100 and 3 variables with degree greater than 1000). To understand the behavior of our algorithm on *cora-1*, we plot the cumulative number of edge updates against the number of vertex updates in Fig. 6.4. We factor out the effect of different parts of `DBRSplash`. We observe that while the work balanced Splash (Sec. 4.1) slightly decreases the cumulative edge updates, the belief scheduling has a more substantially effect.

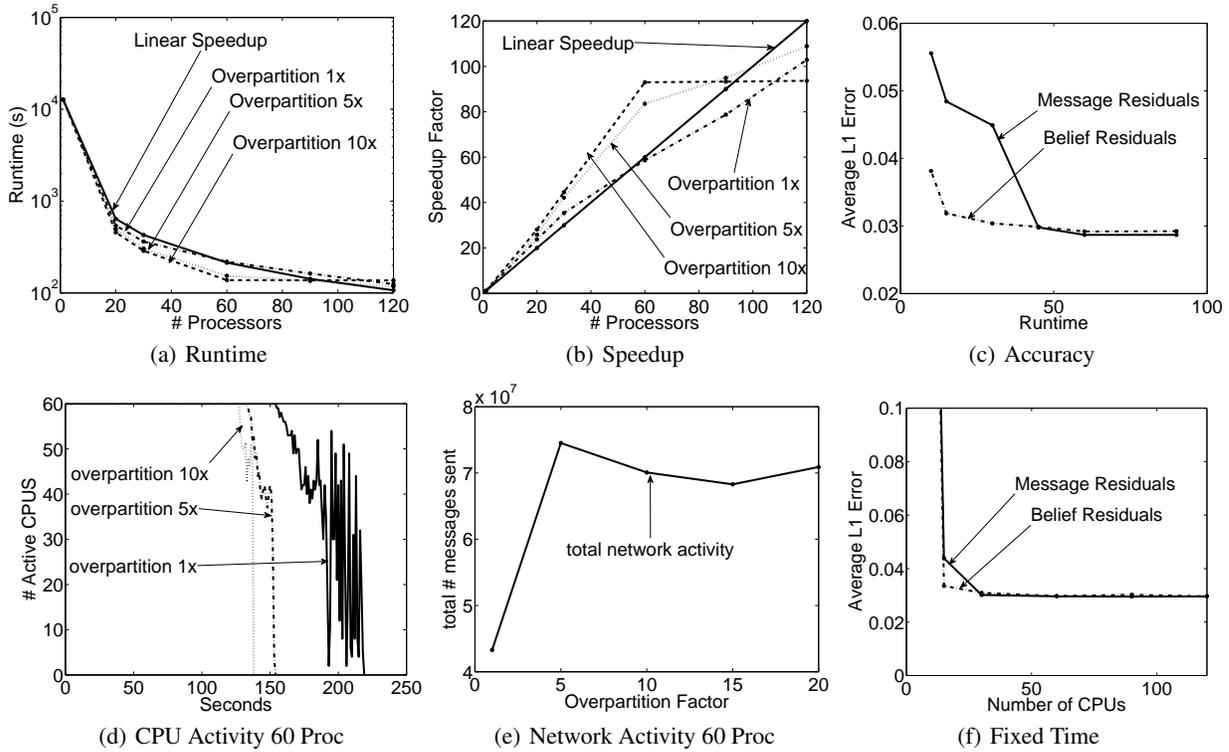

Figure 4: An analysis of DBRSplash on the *uw-systems* MLN. The *uw-systems* MLN is fairly large and therefore amenable to large scale cluster parallelization.

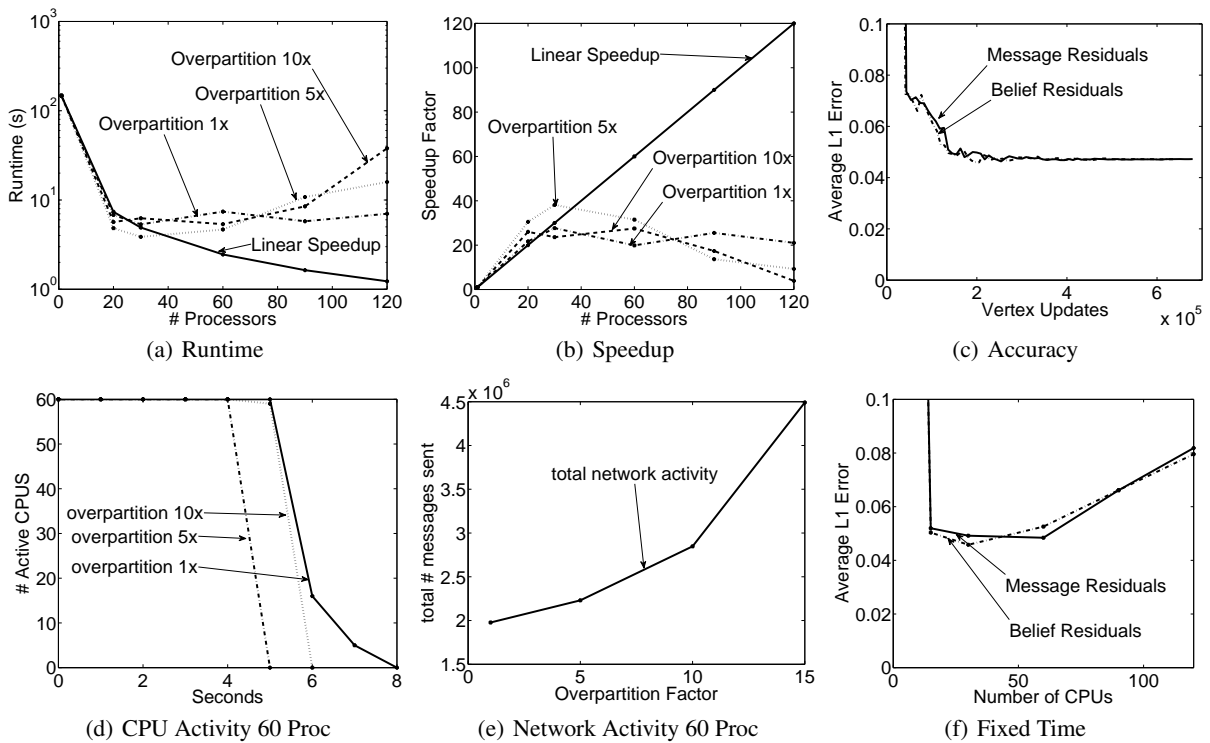

Figure 5: An analysis of DBRSplash on the *uw-languages* MLN. The *uw-languages* MLN is too small for large scale cluster parallelization and therefore demonstrates the failure behaviour of DBRSplash when the graph size is small compared to the number of processors.

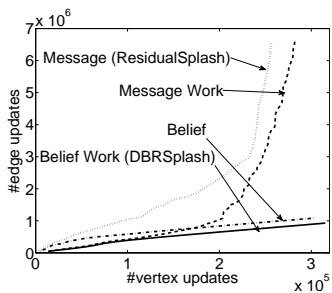

Figure 7: Cumulative edge update counts as a function of time measured in vertex updates for the *cora-1* MLN.

## 7 CONCLUSIONS

We investigated the challenges involved in efficient distributed Belief Propagation on large factor graphs. We identified two primary challenges: efficient state partitioning and scheduling in complex irregular graphs.

To address the problem of state partitioning, we reduced the allocation of factors and messages and computation to graph cuts with edge and vertex weights. While estimating the weights exactly requires knowing the update scheduling, we showed that uninformed cuts perform reasonably well in practice. Because uniformed cuts tend to have lower communication costs and greater imbalance than informed cuts, we proposed over-partitioning to improve balance at the expense of increased communication costs. We found that over-partitioning can reduce the overall running time as long as the communication costs do not dominate.

To support the distributed memory setting and to improve performance on complex irregular graphs we proposed a new scheduling that addresses the limitations in `ResidualSplash` scheduling while retaining the parallel optimality property. Using a distributed collection of queues we decouple the scheduling across processors. By using fixed work sized Splash operations we ensure that high degree vertices are not updated disproportionately often. By switching from message based scheduling to belief based scheduling, we ensure more uniform convergence in the belief estimates. Experimentally, these changes resulted in improved performance, enabling rapid, accurate convergence on graphs which otherwise were intractable using previous belief propagation based techniques.

We tested our new algorithm, `DBRSplash`, on a cluster of 120 processors and found linear to super-linear performance gains on large factor graphs. For small factor graphs which run in minutes on a single processor, we obtained linear speedups only when using up to 20 processors. In conclusion we proposed an efficient parallel distributed algorithm, `DBRSplash`, which performs optimally on large factor graphs and demonstrates the potential capability of efficient parallel algorithms for the future of AI.

### Acknowledgements

This work is supported by ONR Young Investigator Program grant N00014-08-1-0752, the ARO under MURI W911NF0810242, DARPA IPTO FA8750-09-1-0141, the NSF under grants NeTS-NOSS and CNS-0625518 and Joseph Gonzalez is supported by the AT&T Labs Fellowship. We thank Intel Research for cluster time.